\documentclass[11pt,a4paper]{article}
\usepackage[hyperref]{acl2021}
\usepackage{times}
\usepackage{latexsym}

\usepackage{booktabs}
\usepackage{makecell} 
\usepackage{multirow}
\usepackage{multicol}
\usepackage[ruled]{algorithm2e}
\DontPrintSemicolon
\usepackage{adjustbox}
\usepackage{float}
\usepackage{caption}
\usepackage{subcaption}
\usepackage{graphics}
\newsavebox{\tempbox}
\usepackage{amsmath,amsthm,amssymb}
\usepackage{mathtools}
\usepackage{xspace}

\usepackage{ulem}

\makeatletter
\newcommand{\mypm}{\mathbin{\mathpalette\@mypm\relax}}
\newcommand{\@mypm}[2]{\ooalign{%
  \raisebox{.1\height}{$#1+$}\cr
  \smash{\raisebox{-.6\height}{$#1-$}}\cr}}
\makeatother

\DeclarePairedDelimiterX{\infdivx}[2]{(}{)}{%
  #1\;\delimsize|\delimsize|\;#2%
}

\makeatletter
\newcommand{\removelatexerror}{\let\@latex@error\@gobble}
\makeatother

\usepackage{microtype}

\aclfinalcopy 

\title{MATE-KD: Masked Adversarial TExt, a Companion to Knowledge Distillation}

\author{
 Ahmad Rashid\textsuperscript{1}\thanks{$\;\;$Equal Contribution} , Vasileios Lioutas\textsuperscript{2}$^*$\thanks{$\;\;$Work done during an internship at Huawei Noah’s Ark Lab.} , Mehdi Rezagholizadeh\textsuperscript{1} \\
 \textsuperscript{1}Huawei Noah’s Ark Lab, 
 \textsuperscript{2}University of British Columbia \\
 \texttt{\normalsize ahmad.rashid@huawei.com,}
 \texttt{\normalsize contact@vlioutas.com,} \\  \texttt{\normalsize mehdi.rezagholizadeh@huawei.com}
}

\date{}

\begin{document}
\maketitle
\begin{abstract}
The advent of large pre-trained language models has given rise to rapid progress in the field of Natural Language Processing (NLP). While the performance of these models on standard benchmarks has scaled with size, compression techniques such as knowledge distillation have been key in making them practical. We present, MATE-KD, a novel text-based adversarial training algorithm which improves the performance of knowledge distillation. MATE-KD first trains a masked language model based generator to perturb text by maximizing the divergence between teacher and student logits. Then using knowledge distillation a student is trained on both the original and the perturbed training samples. We evaluate our algorithm, using BERT-based models, on the GLUE benchmark and demonstrate that MATE-KD outperforms competitive adversarial learning and data augmentation baselines. On the GLUE test set our 6 layer RoBERTa based model outperforms $\text{BERT}_{\text{LARGE}}$.
\end{abstract}

\section{Introduction}

Transformers~\cite{vaswani2017attention} and transformer-based Pre-trained Language Models (PLMs)~\cite{devlin-etal-2019-bert} are ubiquitous in applications of NLP. Since transformers are non-autoregressive, they are highly parallelizable and their performance scales well with an increase in model parameters and data. Increasing model parameters depends on availability of computational resources and PLMs are typically trained on unlabeled data which is cheaper to obtain.


Recently, the trillion parameter mark has been breached for PLMs~\cite{fedus2021switch} amid serious environmental concerns~\cite{strubell2019energy}. However, without a change in our current training paradigm , training larger models may be unavoidable~\cite{li2020train}. In order to deploy these models for practical applications such as for virtual personal assistants, recommendation systems, e-commerce platforms etc. model compression is necessary.

Knowledge Distillation (KD)~\cite{bucilua2006model,hinton2015distilling} is a simple, yet powerful knowledge transfer algorithm which is used for compression~\cite{jiao2019tinybert,sanh2019distilbert}, ensembling~\cite{hinton2015distilling} and multi-task learning~\cite{clark2019bam}. In NLP, KD for compression has received renewed interest in the last few years. It is one of most widely researched algorithms for the compression of transformer-based PLMs~\cite{rogers2020primer}. 

One key feature which makes KD attractive is that it only requires access to the teacher's output or logits and not the weights themselves. Therefore if a trillion parameter model resides on the cloud, an API level access to the teacher's output is sufficient for KD. Consequently, the algorithm is architecture agnostic i.e. it can work for any deep learning model and the student can be a different model from the teacher. 

Recent works on KD for transfer learning with PLMs extend the algorithm in two main directions. The first is towards ``model" distillation~\cite{sun2019patient, wang2020minilm,jiao2019tinybert} i.e. distilling the intermediate weights such as the attention weights or the intermediate layer output of transformers. The second direction is towards curriculum-based or progressive KD~\cite{sun2020mobilebert, mirzadeh2019improved} where the student learns one layer at a time or from an intermediary teacher, known as a teacher assistant. While these works have shown accuracy gains over standard KD, they have come at the cost of architectural assumptions, least of them a common architecture between student and teacher, and greater access to teacher parameters and intermediate outputs. Another issue is that the decision to distill one teacher layer and to skip another is arbitrary. Still the teacher typically demonstrates better generalization  

We are interested in KD for model compression and study the use of adversarial training~\cite{goodfellow2014explaining} to improve student accuracy using just the logits of the teacher as in standard KD. Specifically,  our work makes the following contributions:

\begin{itemize}
    \item We present a text-based adversarial algorithm, MATE-KD, which increases the accuracy of the student model using KD.
    \item Our algorithm only requires access to the teacher's logits and thus keeps the teacher and student architecture independent.
    \item We evaluate our algorithm on the GLUE \cite{wang-etal-2018-glue} benchmark and demonstrate improvement over competitive baselines.
    \item On the GLUE test set we achieve a score of 80.9, which is higher than $\text{BERT}_{\text{LARGE}}$
    \item We also demonstrate improvement on out-of-domain (OOD) evaluation.
\end{itemize}

\section{Related Work}
\subsection{Knowledge Distillation}
We can summarize the knowledge distillation loss, $\mathcal{L}$, as following: 
\begin{equation}
    \begin{split}
        & \mathcal{L}_{CE} = \mathcal{H}_{CE}\big( y , S(X)) \big) \\ 
        & \mathcal{L}_{KD} = \mathcal{T}^2 D_{KL}\Big(\sigma(\frac{z_t(X)}{\mathcal{T}}), \sigma(\frac{z_s(X)}{\mathcal{T}})\Big) \\
        & \mathcal{L} = (1-\lambda) \mathcal{L}_{CE} + \lambda \mathcal{L}_{KD} 
    \end{split}
    \label{eq:KD}
\end{equation}
where $\mathcal{H}_{CE}$ represents the cross entropy between the true label $y$ and the student network prediction $S(X)$ for a given input $X$, $D_{KL}$ is the KL divergence between the teacher and student predictions softened using the temperature parameter $\mathcal{T}$, $z(X)$ is the network output before the softmax layer (logits), and $\sigma (.)$ indicates the softmax function. The term $\lambda$ in the above equation is a hyper-parameter which controls the amount of contribution from the cross entropy and KD loss. 

Patient KD~\cite{sun2019patient} introduces an additional loss to KD which distills the intermediate layer information onto the student network. Due to a difference in the number of student and teacher layers they propose either skipping alternate layers or distilling only the last few layers. TinyBERT~\cite{jiao2019tinybert} applies embedding distillation and intermediate layer distillation which includes hidden state distillation and attention weight distillation. Although it achieves strong results on the GLUE benchmark this approach is infeasible for very large teachers. MiniLM~\cite{wang2020minilm} proposed an interesting alternative whereby they distill the key, query and value matrices of the final layer of the teacher.

\subsection{Adversarial Training}
Adversarial examples are small perturbations to training samples indistinguishable to humans but enough to produce misclassifications by a trained neural network. \citet{goodfellow2014explaining} showed that adding these examples to the training set can make a neural network model robust to perturbations. \citet{miyato2016adversarial} adapt adversarial training to text classification and improve performance on a few supervised and semi-supervised text classification tasks. 

In NLP, adversarial training has surprisingly been shown to  improve generalization as well~\cite{cheng2019robust,zhu2019freelb}.  \citet{cheng2019robust} study machine translation and propose making the model robust to both source and target perturbation, generated by swapping the word embedding of a word with that of its synonym. They model small perturbations by considering word swaps which cause the smallest increase in loss gradient. They achieve a higher BLEU score on Chinese-English and English-German translation compared to the baseline. 

\citet{zhu2019freelb} propose a novel adversarial training algorithm, FreeLB, to make gradient based adversarial training efficient by updating both embedding perturbations and model parameters simultaneously during the backward pass of training. They show improvements on multiple language models on the GLUE  benchmark. Embedding based perturbations are attractive because they produce stronger adversaries~\cite{zhu2019freelb} and keep the system end-to-end differentiable as embeddings are continuous. The salient features of  adversarial training for NLP are a) a \textit{minimax} formulation where the adversarial examples are generated to maximize a loss function and the model is trained to minimize the loss the function and b) a way of keeping the perturbations small such as a norm-bound on the gradient~\cite{zhu2019freelb} or replacing words by their synonyms~\cite{cheng2019robust}.  

If these algorithms are adapted to KD one key challenge is the embedding mismatch between the teacher and student. Even if the embedding size is the same, the student embedding needs to be frozen to match the teacher embedding and freezing embeddings typically leads to lower performance. If we adapt adversarial training to KD, one key advantage is that access to the teacher distribution relaxes the requirement of generating label preserving perturbations. These considerations have prompted us to design an adversarial algorithm where we perturb the actual text instead of the embedding. We will demonstrate in our methodology how we address the differentiability question.

\subsection{Data Augmentation}

One of the first works on BERT compression~\cite{tang2019distilling} used KD and proposed data augmentation using heuristics such as Part-Of-Speech guided word replacement. They demonstrated improvement on three GLUE tasks. One limitation of this approach is that the heuristics are task specific. \citep{jiao2019tinybert} present an ablation study in their work whereby they demonstrate a strong contribution of data augmentation to their KD algorithm performance. They augment the data by randomly selecting a few words of  a training sentence and replacing them with words with the closest embedding under cosine distance. Our adversarial learning algorithm can be interpreted as a data augmentation algorithm but instead of a heuristic approach we propose a principled end-to-end differentiable augmentation method based on adversarial learning.

\citep{khashabi2020more} presented a data augmentation technique for question answering whereby they took seed questions and asked humans to perturb only a few tokens to generate new ones. The human annotators could modify the label if needed. They demonstrated improved generalization and robustness with the augmented data. We will demonstrate that our algorithm is built with similar principles but does not require humans in the loop. Instead of human annotators to modify the labels we use the teacher.

\section{Methodology}

\begin{figure}[h]
    \centering
    \includegraphics[scale=0.33]{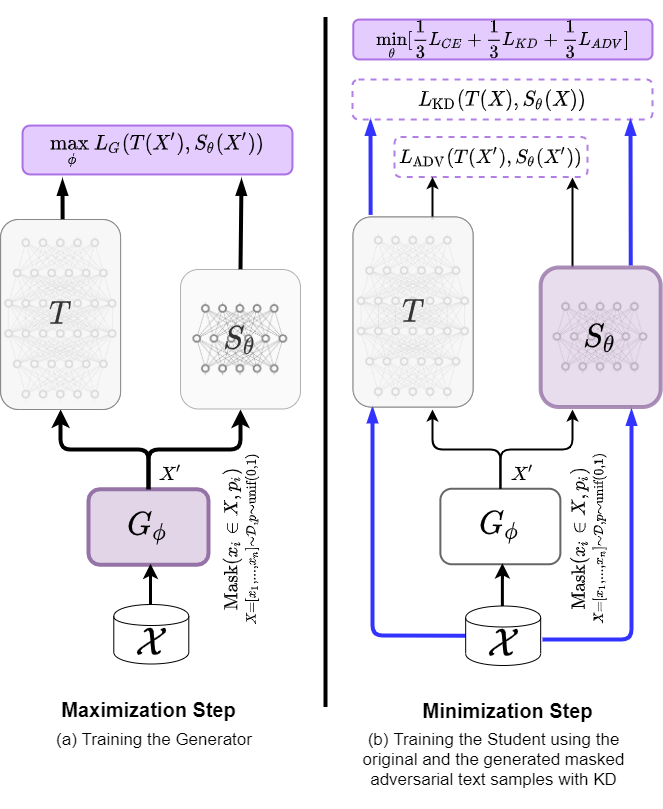}
    \caption{Illustration of the maximization and minimization steps of MATE-KD }
    \label{fig:mate-kd}
\end{figure}

We propose an algorithm that involves co-training and deploy an adversarial text generator while training a student network using KD. Figure~\ref{fig:mate-kd} gives an illustration of our architecture.

\subsection{Generator}
The text generator is simply a pre-trained masked language model which is trained to perturb training samples adversarially. We can frame our technique in a \textit{minimax} regime such that in the maximization step of each iteration, we feed the generator with a training sample with few of the tokens replaced by masks. We fix the rest of the sentence and replace the masked tokens with the generator output to construct a pseudo training sample $X'$. This pseudo sample is fed to both the teacher and the student models and the generator is trained to maximize the divergence between the teacher and the student. We present an example of the masked generation process in Figure~\ref{fig:mask}. The student is trained during the minimization step. 


\subsection{Maximization Step}
The generator is trained to generate pseudo samples by maximizing the following loss function:    
\begin{equation}
    \begin{split}
      &  \max_\phi \mathcal{L}_G (\phi) = \\ 
      & D_{KL}\Big( T\big( G_\phi(X^m)\big), S_{\theta}\big(G_\phi(X^m)\big) \Big), \\ 
    \end{split}
\end{equation}
where $D_{KL}$ is the KL divergence, $G_\phi$(.) is the text generator network with parameters $\phi$, $T(\cdot)$ and $S_{\theta}(\cdot)$ are the teacher and student networks respectively, and $X^m$ is a randomly masked version of the input $X = [x_1, x_2, ..., x_n]$ with $n$ tokens.

\begin{align}
\begin{split}
     & \forall x_i \in X = [x_1,..., x_i, ..., x_n]\sim \mathcal{D},\\
     & x_i^m = \underset{p\sim \text{unif}(0,1)}{\text{Mask}(x_i \in X, p_i)} \\ 
    & =
    \begin{cases}
        x_i,&p_i \ge \rho\\
        <\text{mask}>, &\text{o.w.}
    \end{cases}
\end{split}
\end{align}
where $\text{unif}(0,1)$ represents the uniform distribution, and the $\text{Mask(}\cdot\text{)}$ function masks the tokens of inputs sampled from the data distribution $\mathcal{D}$ with the probability of  $\rho$. The term $\rho$ can be treated as a hyper-parameter in our technique. In summary, for each training sample, we randomly mask some tokens according to the samples derived from the uniform distribution and the threshold value of $\rho$. 

Then in the forward pass, the masked sample, $X^m$, is fed to the generator to obtain the output pseudo text based on the generator predictions of the mask tokens. The generator needs to output a one-hot representation but using an \textit{argmax} inside the generator would lead to non-differentiability. Instead we apply the Gumbel-Softmax~\cite{jang2016categorical}, which, is an approximation to sampling from the \textit{argmax}. Using the straight through estimator~\cite{bengio2013estimating} we can still apply \textit{argmax} in the forward pass and can obtain text, $X'$ from the network outputs:


\begin{figure}[t]
    \centering
    \includegraphics[scale=0.33]{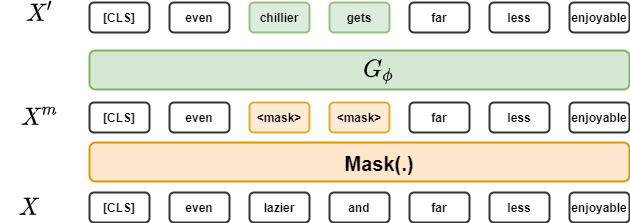}
    \caption{This figure illustrates how a training sample will be randomly masked and then fed to the text generator $G_\phi$ to get the pseudo training sample.}
    \label{fig:mask}
\end{figure}
\begin{equation}
      X' = \underset{\text{FORWARD}}{G_\phi(X^m)} = \text{argmax}\big(\sigma_{\text{Gumbel}}(z_\phi(X^m)\big) 
\end{equation}
where 
\begin{align}
    & \sigma_{\text{Gumbel}}(z_i) = 
      & \frac{\exp{\Big(\big(\log (z_i) +g_i \big)/\tau\Big) }}{\Sigma_{j=1}^K \exp{\Big(\big(\log (z_j) +g_j \big)/\tau \Big) }}
\end{align}
$g_i\sim \text{Gumbel}(0,1)$ and $z_\phi(.)$ returns the logits produced by the generator for a given input.

In the backward pass the generator simply applies the gradients from the Gumbel-Softmax without the \textit{argmax} :

\begin{equation}
      \underset{\text{BACKWARD}}{G_\phi(X^m)} = \sigma_{\text{Gumbel}}(z_\phi(X^m) )
\end{equation}


\subsection{Minimization Step}
In the minimization step, the student network is trained such as to minimize the gap between the teacher and student predictions  and match the hard labels from the training data by minimizing the following loss equation:
\begin{equation}
    \begin{split}
        & \min_\theta \mathcal{L}_\text{MATE-KD}(\theta) = \\
        & \frac{1}{3} \mathcal{L}_{CE}(\theta) + \frac{1}{3} \mathcal{L}_{KD}(\theta) + \frac{1}{3} \mathcal{L}_{ADV}(\theta)
    \end{split}
    \label{eq:loss}
\end{equation}
where
\begin{equation}
    \mathcal{L}_{ADV}(\theta) = D_{KL}\Big( T(X') , S_{\theta}(X') \Big)
\end{equation}

In Equation \ref{eq:loss}, the terms ${L}_{KD}$ and ${L}_{CE}$ are the same as Equation~\ref{eq:KD}, ${L}_{KD}(\theta)$ and ${L}_{ADV}(\theta)$ are used to match the student with the teacher, and ${L}_{CE}(\theta)$ is used for the student to follow the ground-truth labels $y$. 

Bear in mind that our $\mathcal{L}_\text{MATE-KD}(\theta)$ loss is different from the regular KD loss in two aspects: first, it has the additional adversarial loss, $\mathcal{L}_{ADV}$ to minimize the gap between the predictions of the student and the teacher with respect to the generated masked adversarial text samples, $X'$, in the maximization step; second, we do not have the weight term $\lambda$ form KD in our technique any more (i.e. we consider equal weights for the three loss terms in $\mathcal{L}_\text{MATE-KD}$).


\subsection{Rationale Behind the Masked Adversarial Text Generation for KD}

The rationale behind generating partially masked adversarial texts instead of generating adversarial texts from scratch (that is equivalent to masking the input of the text generator entirely) is three-fold: 
\begin{enumerate}
    \item Partial masking is able to generate more realistic sentences compared to generating them from scratch when trained only to increase teacher and student divergence. We present a few generated sentences in section~\ref{sec:gen_sentences} 
    \item Generating text from scratch increases the chance of generating OOD data. Feeding OOD data to the KD algorithm leads to matching the teacher and student functions across input domains that the teacher is not trained on.
    \item By masking and changing only a few tokens of the original text, we constrain the amount of perturbation as is required for adversarial training.
\end{enumerate}

In our MATE-KD technique, we can tweak the $\rho$ to control our divergence from the data distribution and find the sweet spot which gives rise to maximum improvement for KD. We also present an ablation on the effect of this parameter on downstream performance in section~\ref{sec:sensitivity}.   


\section{Experiments}
We evaluated MATE-KD on all nine datasets of the General Language Understanding Evaluation (GLUE) \citep{wang-etal-2018-glue} benchmark which include classification and regression. These datasets can be broadly divided into 3 family of problems. Single set tasks which include linguistic acceptability (CoLA) and sentiment analysis (SST-2). Similarity and paraphrasing tasks which include paraphrasing (MRPC and QQP) and a regression task (STS-B). Inference tasks which include Natural Language Inference (MNLI, WNLI, RTE) and Question Answering (QNLI). 

\begin{table*}[t]
\footnotesize
    \centering
    \begin{tabular}{lccccccccc}
        \toprule
        Method & CoLA & SST-2 & MRPC & STS-B & QQP & MNLI & QNLI & RTE & Score  \\
        \midrule
        $\text{RoBERTa}_\text{Large}$ (teacher) & 68.1 & 96.4 & 91.9 & 92.3 & 91.5 & 90.2 & 94.6 & 86.3 & 85.28 \\
        \midrule
        DistilRoBERTa (student) & 56.6 & 92.7 & 89.5 & 87.2 & 90.8 & 84.1 & 91.3 & 65.7 & 78.78 \\
        Student + FreeLB & 58.1 & 93.1 & 90.1 & 88.8 & 90.9 & 84.0 & 91.0 & 67.8 & 80.01 \\
        Student + FreeLB + KD & 58.1 & 93.2 & 90.5 & 88.6 & 91.2 & 83.7 & 90.8 & 68.2 & 80.06 \\
        Student + KD & 60.9 & 92.5 & 90.2 & 89.0 & 91.6 & 84.1 & 91.3 & 71.1 & 80.77 \\
        Student + TinyBERT Aug + KD & 61.3 & 93.3 & 90.4 & 88.6 & 91.7 & 84.4 & 91.6 & 72.5 & 81.12 \\ 
        \midrule
        Student + MATE-KD (Ours) & 65.9 & 94.1 & 91.9 & 90.4 & 91.9 & 85.8 & 92.5 & 75.0 & \textbf{82.64} \\
         \bottomrule
    \end{tabular}
    \caption{Dev Set results for the GLUE benchmark. The score for the WNLI task is 56.3 for all models.}
    \label{tab:dev_results}
\end{table*}

\subsection{Experimental Setup}
We evaluate our algorithm on two different setups. On the first the teacher model is $\text{RoBERTa}_{\text{LARGE}}$ \cite{liu2020roberta} and the student is initialized with the weights of DistillRoBERTa \cite{sanh2019distilbert}. $\text{RoBERTa}_{\text{LARGE}}$ consists of 24 layers with a hidden dimension of 1024 and 16 attention heads and a total of 355 million parameters. We use the pre-trained model from Huggingface \cite{Wolf2019HuggingFacesTS}. The student consists of 6 layers, 768 hidden dimension, 8 attention heads and 82 million parameters. Both models have a vocabulary size of 50,265 extracted using the Byte Pair Encoding (BPE) \citep{sennrich-etal-2016-neural} tokenization method.

On our second setup the teacher model is $\text{BERT}_{\text{BASE}}$ \cite{devlin-etal-2019-bert} and the student model is initialized with the weights of DistilBERT which consists of 6 layers with a hidden dimension of 768 and 8 attention heads. The pre-trained models are taken from the authors release. The teacher and the student are 110M and 66M parameters respectively with a vocabulary size of 30,522 extracted using BPE. 

\paragraph{Hyper-parameters\footnote{We will release the code upon the acceptance of the paper.}} We fine-tuned the distill RoBERTa-based student model for 30 epochs and picked the best checkpoint that gave the highest score on the dev set of GLUE. These checkpoints was also used for the test set submission. For each task, we selected the best fine-tuning learning rate among 5e-5, 4e-5, 3e-5, and 2e-5 values. We used the AdamW \citep{loshchilov2017decoupled} optimizer with the default values. In addition, we used a linear decay learning rate scheduler with no warmup steps. We set the masking probability $p$ to be 0.3. Additionally, we set the value $n_G$ to 10 and $n_S$ to 100. For the distillBERT student model we used the same hyper-parameters that we use for the distillRoBERTa student. 

\paragraph{Hardware Details} We trained all models using a single NVIDIA V100 GPU. We used mixed-precision training \citep{micikevicius2018mixed} to expedite the training procedure. All experiments were run using the PyTorch\footnote{\url{https://pytorch.org/}} framework.

\subsection{Results}

Table~\ref{tab:dev_results} presents the results of MATE-KD on the GLUE dev set. Even though the datasets have different evaulation metrics, we present the average of all scores as well, which is used to rank the submissions to GLUE. Our first baseline is the fine-tuned DistilRoBERTa and then we compare with KD, FreeLB, FreeLB plus KD, and TinyBERT~\cite{jiao2019tinybert} Augmentation plus KD.

We observe that FreeLB \cite{zhu2019freelb} significantly improves the fine-tuned student by around 1.2 points on average. However, when we apply both FreeLB + KD we don't see any further improvement whereas applying KD alone improves the score by about 2 points. This is so because FreeLB relies on the model (student) output rather than the teacher output to generate adversarial perturbation and therefore cannot benefit from KD. As previously discussed, FreeLB relies on embedding perturbation and in order to generate the teacher output on the perturbed student both the embeddings need to be tied together which is infeasible due to size and training requirements. 


We also compared against the data augmentation algorithm of TinyBERT. We ran their code to generate the augmented data offline. Although they augment the data about 20 times depending on the GLUE task we observed poor results if we use all this data to fine-tune with KD. We only generated 1x augmented data and saw an average improvement of 0.35 score over KD. MATE-KD achieves the best result among the student models on all the GLUE tasks and achieves an average improvement of 1.87 over just KD. We also generated the same number of adversarial samples as the training data.

\begin{table*}[t]
\footnotesize
    \centering
    \begin{tabular}{lccccccccc}
        \toprule
        Model (Param.) & CoLA & SST-2 & MRPC & STS-B & QQP & MNLI-m/mm & QNLI & RTE & Score  \\
        \midrule
        TinyBERT (66M) & 51.1 & 93.1 & 87.3/82.6 & 85.0/83.7 & 71.6/89.1 & 84.6/83.2 & 90.4 & 70.0 & 78.1 \\
        $\text{BERT}_\text{BASE}$ (110M) & 52.1 & 93.5 & 88.9/84.8 &  87.1/85.8 & 71.2/89.2 & 84.6/83.4 & 90.5 & 66.4 & 78.3 \\
        MobileBERT (66M) & 51.1 & 92.6 & 88.8/84.5 & 86.2/84.8 & 70.5/88.3 & 84.3/83.4 & 91.6 & 70.4 & 78.5 \\
        DistilRoB. + KD (82M) & 54.3 & 93.1 & 86.0/80.8 & 85.7/84.9 & 71.9/89.5 & 83.6/82.9 & 90.8 & 74.1 & 78.9 \\ 
        $\text{BERT}_\text{LARGE}$ (340M) & 60.5 & 94.9 & 89.3/85.4 &  87.6/86.5 & 72.1/89.3 & 86.7/85.9 & 92.7 & 70.1 & 80.5 \\
        \midrule
        \textit{MATE-KD} (82M) & 56.0 & 94.9 & 91.7/88.7 & 88.3/87.7 & 72.6/89.7 & 85.5/84.8 & 92.1 & 75.0 & \textbf{80.9} \\
        \bottomrule
    \end{tabular}
    \caption{Leaderboard test results of experiments on GLUE tasks. The score for the WNLI task is 65.1 for all models.}
    \label{tab:test_res}
\end{table*}

We present the results on the test set of GLUE on Table~\ref{tab:test_res}. We list the number of parameters for each model. The results of $\text{BERT}_\text{BASE}$, $\text{BERT}_\text{LARGE}$ \cite{devlin-etal-2019-bert}, TinyBERT and MobileBERT \cite{sun2020mobilebert} are taken from the GLUE leaderboard\footnote{\url{https://gluebenchmark.com/leaderboard}}. The KD models have $\text{RoBERTa}_\text{Large}$, fine-tuned without ensembling, as the teacher. 

TinyBERT and MobileBERT are the current state-of-the-art models when it comes to 6 layer BERT-based models on the GLUE leaderboard. We include them in this comparison although their teacher is $\text{BERT}_\text{BASE}$ as opposed to $\text{RoBERTa}_\text{Large}$. We make the case that one reason we can train with a larger and more powerful teacher is that we only require logits of the teacher while training. Most of the works in the literature proposing intermediate layer distillation ~\cite{jiao2019tinybert, sun2020mobilebert, sun2019patient} are trained on 12 layer BERT teachers. As PLMs get bigger in size feasible approaches to KD will involve algorithms which rely on only minimal access to teachers.  

We apply a standard trick to boost the performance of SST-B and RTE, i.e., we initialize these models with the trained checkpoint of MNLI \cite{liu2020roberta}. This was not done for the dev results. The WNLI score is the same for all the models and although, not displayed on the table is part of the average score. We make a few observations from this table. Firstly, using KD a student with a powerful teacher can overcome a significant difference in parameters between competitive models. Secondly, our algorithm significantly improves KD with an average 2 point increase on the unseen GLUE testset. Our model is able to achieve state-of-the-art results for a 6 layer BERT-based model on the GLUE leaderboard.

\begin{table*}[t]
\footnotesize
    \centering
    \begin{tabular}{lccccccccc}
        \toprule
        Method & CoLA & SST-2 & MRPC & STS-B & QQP & MNLI & QNLI & RTE & Score  \\
        \midrule
        $\text{BERT}_\text{BASE}$ (teacher) & 59.5 & 93.1 & 86.7 & 88.4 & 91.0 & 84.6 & 91.5 & 68.2 & 82.9 \\
        \midrule
        $\text{DistilBERT}_\text{BASE}$ (student) & 51.3 & 91.3 & 87.5 & 86.9 & 88.5 & 82.1 & 89.2 & 59.9 & 79.6 \\
        Student + TinyBERT Aug. + KD & 55.2 & 91.9 & 87.0 & 87.8 & 89.5 & 82.1 & 89.7  & 68.6 & 81.5 \\ 
        \midrule
        Student + MATE-KD (Ours) & 60.4 & 92.2 & 88.0 & 88.5 & 91.4 & 84.5 & 91.2 & 70.0  & \textbf{83.3} \\
         \bottomrule
    \end{tabular}
    \caption{Dev results on the GLUE benchmark using $\text{BERT}_\text{BASE}$ as the teacher model and $\text{DistilBERT}_\text{BASE}$ as the student model.}
    \label{tab:dev_results_2}
\end{table*}

We also evaluate our algorithm using $\text{BERT}_\text{BASE}$ as teacher and DistilBERT as student on GLUE benchmark. WNLI results are not reported. We compare against the teacher, student and KD plus TinyBERT augmentation. Here, remarkably MATE-KD can beat the teacher performance on average. On the two large datasets in GLUE, QQP and MNLI, we beat and match the teach performance respectively.

We observe that Mate-KD outperforms its competitors when both the teacher is twice the size and four times the size of the student. This may be because the algorithm generates adversarial examples based on the teacher's distribution. A well designed adversarial algorithm can help us probe parts of the teacher's distribution not spanned by the training data leading to better generalization.

\subsection{OOD Evaluation}

It has been shown that strong NLU models tend to learn spurious, surface level patterns from the dataset \cite{poliak2018hypothesis,gururangan2018annotation} and may perform poorly on carefully constructed OOD datasets. In Table~\ref{tab:ood} we present the evaluation of Mate-KD (RoBERTa-based) trained on MNLI and QQP  on the HANS~\cite{mccoy2019right} and the PAWS~\cite{zhang2019paws} evaluation sets respectively.

\begin{table}[!th]

	\begin{center}
		\begin{tabular}{l|c|c}
			\toprule
			Model & HANS & PAWS \\
			 \midrule
			DistilRoBERTa & 58.9 & 36.5  \\
			Mate-KD & 66.6 & 38.3  \\
			\bottomrule
		\end{tabular}
	\end{center}
	\caption{Model Performance on OOD evaluation sets HANS and PAWS for MNLI and QQP respectively}
	\label{tab:ood}

\end{table}

We use the same model checkpoint as the one presented in table~\ref{tab:dev_results} and compare against DistilRoBERTa. We observe that MATE-KD improves the baseline performance on both evaluation datasets. The performance increase on HANS is larger. We can conclude that the algorithm improvements are not due to learning spurious correlations and biases in the dataset. 

\subsection{Ablation Study}

Table~\ref{tab:abl} presents the contribution of the generator and adversarial learning to MATE-KD. We first present the result of MATE-KD on 4 diverse datasets from GLUE and compare against the effect of removing the adversarial training and then the generator altogether. When we remove adversarial training we essentially remove the maximization step and do not train the generator. The generator in this setting is a pre-trained masked language model. In the minimization step we still generate pseudo samples and apply all the losses. The setting where we remove the generator is akin to simple KD.

We observe that across all four datasets both a simple generator without the maximization step and adversarial training contribute to the performance of MATE-KD.   

\begin{table}[!th]
    \centering
    \begin{tabular}{lcccc}
    \toprule
    Model & SST-2 & QQP & MNLI & RTE   \\
    \midrule
    MATE-KD  & \textbf{94.1} & \textbf{91.9} & \textbf{85.8} & \textbf{75.0} \\
    \indent - Adv train  & 93.1 & 91.8 & 85.3 & 74.0 \\
    \indent - Generator & 92.5 & 91.6 & 84.1 & 71.1 \\
    \bottomrule
    \end{tabular}
\caption{The ablation of MATE-KD on four datasets from the GLUE benchmark. We present the result of MATE-KD, algorithm without training generator and algorithm without generator. Results are on the dev set.}
\label{tab:abl}
\end{table}

\begin{table*}[!htbp]
\centering
\begin{tabular}{c|c} 
\textbf{Original} & \textbf{Generated} \\ \hline
\textcolor{cyan}{the} new insomnia \textcolor{cyan}{is} a surprisingly \textcolor{cyan}{faithful} & \textcolor{green}{sinister} new insomnia \textcolor{green}{shows} a surprisingly \textcolor{green}{terrible} \\ 
remake of its \textcolor{cyan}{chilly} predecessor, and & remake of its \textcolor{green}{hilarious} predecessor, and \\ \hline
beautifully \textcolor{cyan}{shot}, delicately scored \textcolor{cyan}{and} &  beautifully \textcolor{green}{sublime}, delicately scored\textcolor{green}{,} \\ 
powered by \textcolor{cyan}{a set} of heartfelt performances & powered by \textcolor{green}{great dozens} of heartfelt performances \\ \hline
a \textcolor{cyan}{perfectly} pleasant if slightly pokey comedy & a \textcolor{green}{10} pleasant if slightly pokey comedy \\ \hline
\textcolor{cyan}{that} appeals \textcolor{cyan}{to} me & \textcolor{green}{Federal} appeals \textcolor{green}{punished} me \\ \hline
good news \textcolor{cyan}{to} anyone who's fallen under & good news \textcolor{green}{for} anyone who's fallen under  \\ the sweet, melancholy spell of this & the sweet, melancholy spell of this \\ unique director's previous \textcolor{cyan}{films} & unique director's previous \textcolor{green}{mistakes} \\

\bottomrule

\end{tabular}
\caption{Examples of original and adversarially generated samples during training for the SST-2 dataset}
\label{tab:gen_sentences}
\end{table*}

\subsection{Sensitivity Analysis}
\label{sec:sensitivity}

\begin{table}[]
    \centering
    \begin{tabular}{l|ccccc}
        \toprule
        \multirow{2}{*}{Task} & \multicolumn{5}{c}{$p\;$ Hyperparameter}  \\
        \cline{2-6}
        & 10\% & 20\% & 30\% & 40\% & 50 \% \\
        \midrule
        MNLI & 85.4 & 85.5 & 85.8 & 84.7 & 84.6 \\
        RTE & 74.0 & 74.8 & 75.0 & 75.4 & 74.6 \\
        \bottomrule
    \end{tabular}
    \caption{$\rho$ value sensitivity analysis on two GLUE tasks.}
    \label{tab:p_sensitivity}
\end{table}

Our algorithm does not require the loss interpolation weight of KD but instead relies on one additional parameter, $\rho$, which is the probability of masking a given token. We present the effect of changing $\rho$ in Table~\ref{tab:p_sensitivity} on MNLI and RTE dev set results fixing all the other hyper-parameters. We selected MNLI and RTE because they are part of Natural Language Inference which is one of the hardest tasks on GLUE. Moreover, on the RoBERTa experiments we see the largest drop on student scores for these two datasets. We can observer that for MNLI the best result is for 30\% followed by 20\% and for RTE the best choice is 40\% followed by 30\%. This corresponds to the heuristic based data augmentation works where they typically modify tokens with a 30\% to 40\% probability. We set this parameter to 30\% for all the experiments and did not tune this for each dataset or each architecture.

\subsection{Generated Samples}
\label{sec:gen_sentences}
We present a few selected samples that our generator produced during training for the SST-2 dataset on table~\ref{tab:gen_sentences}. SST-2 is a binary sentiment analysis dataset. The data consist of movie reviews and is both at the phrase and sentence level. 

We observe that we only modify a few tokens in the generated text. However, one of three things happens if the text is semantically plausible. Either the generated sentence keeps the same sentiment as in examples 2 and 3, or it changes the sentiment as in example 1 and 4 or the text has ambiguous sentiment as in example 5. We can use all of these for training since we don't rely on the original label but obtain the teacher's output.

\section{Discussion and Future Work}

We have presented MATE-KD, a novel text-based adversarial training algorithm which improves the student model in KD by generating adversarial examples while accessing the logits of the teacher only. This approach is architecture agnostic and can be easily adapted to other applications of KD such as model ensembling and multi-task learning.

We demonstrate the need for a text-based, rather than embedding perturbation based, adversarial training algorithm for KD. We demonstrate the importance of masking in our algorithm and show that it is vital for improving student performance.

One key theme that we have presented in this work is that as PLM's inevitably increase in size and number of parameters, techniques that rely on access to the various layers and intermediate parameters of the teacher will be more difficult to train. Not only that, the decision of what to distill and where to distill from is sometimes arbitrary. In contrast, algorithms which are well-motivated, and require minimal access to the teacher may learn from more powerful teachers and would be more useful. An example of such an algorithm is the KD algorithm itself.

Future work will consider a) using label information and a measure of semantic quality to filter the generated sentences b) exploring the application of masked based adversarial training to continuous data such as speech and images and c) exploring other applications of KD.

\section*{Impact Statement}

Our research primarily deals with deploying high quality NLP applications to a wide audience around the globe. We contend that these technologies can simplify many of our mundane tasks and free up our time to pursue more pleasurable work. We are well aware that in themselves they are no panacea and any time that they free up may well be exploited by the very companies which distribute these for profit. However, we try to be humble and solve the problem at hand and encourage all the stakeholders to play their part.

\bibliographystyle{acl_natbib}
\bibliography{anthology,acl2021}


\end{document}